\documentclass[letterpaper,10pt,conference]{ieeeconf}

\IEEEoverridecommandlockouts

\usepackage[utf8]{inputenc} 
\usepackage[T1]{fontenc}    
\usepackage{hyperref}       
\usepackage{url}            
\usepackage{booktabs}       
\usepackage{amsfonts}       
\usepackage{nicefrac}       
\usepackage{microtype}      

\usepackage{times}
\usepackage{graphicx} 
\usepackage{caption}
\usepackage{subcaption}
\usepackage{wrapfig}
\usepackage{rotating}

\usepackage{algorithm}
\usepackage{algorithmic}

\usepackage{hyperref}

\usepackage{amsmath}
\usepackage{amssymb}

\usepackage[font=footnotesize]{caption,subcaption}
\usepackage{sidecap}

\usepackage{xspace}

\usepackage{color}

\newcommand{\bs}{\mathbf{s}}
\newcommand{\ba}{\mathbf{a}}
\newcommand{\bA}{\mathbf{A}}

\newcommand{\pushright}[1]{\ifmeasuring@#1\else\omit\hfill$\displaystyle#1$\fi\ignorespaces}

\newcommand{\specialcell}[2][c]{%
  \begin{tabular}[#1]{@{}c@{}}#2\end{tabular}}
  
\DeclareGraphicsExtensions{.png}
\graphicspath{ {./figures/} }

\title{\LARGE \bf
Self-supervised Deep Reinforcement Learning\\with Generalized Computation Graphs for Robot Navigation
}

\author{Gregory Kahn, Adam Villaflor, Bosen Ding, Pieter Abbeel, Sergey Levine\\
Berkeley AI Research (BAIR), University of California, Berkeley%
}

\begin{document}

\maketitle
\thispagestyle{empty}
\pagestyle{empty}


\begin{abstract} 
Enabling robots to autonomously navigate complex environments is essential for real-world deployment. Prior methods approach this problem by having the robot maintain an internal map of the world, and then use a localization and planning method to navigate through the internal map. However, these approaches often include a variety of assumptions, are computationally intensive, and do not learn from failures. In contrast, learning-based methods improve as the robot acts in the environment, but are difficult to deploy in the real-world due to their high sample complexity. To address the need to learn complex policies with few samples, we propose a generalized computation graph that subsumes value-based model-free methods and model-based methods, with specific instantiations interpolating between model-free and model-based. We then instantiate this graph to form a navigation model that learns from raw images and is sample efficient. Our simulated car experiments explore the design decisions of our navigation model, and show our approach outperforms single-step and $N$-step double Q-learning. We also evaluate our approach on a real-world RC car and show it can learn to navigate through a complex indoor environment with a few hours of fully autonomous, self-supervised training. Videos of the experiments and code can be found at \url{github.com/gkahn13/gcg}
\end{abstract} 


\section{Introduction}
\label{sec:intro}


In order to create robots that can autonomously navigate complex and unstructured environments, such as roads, buildings, or forests, we need generalizable perception and control systems that can reason about the outcomes of navigational decisions. Although methods based on geometric reconstruction and mapping have proven effective in a range of navigation and collision avoidance domains~\cite{thorpe1988vision,urmson2008autonomous,shen2011autonomous}, they impose considerable computational overhead~\cite{olson2008phd} and often include a variety of assumptions~\cite{fuentes2015visual} that may not hold in practice, such as static environments or absence of transparent objects. In this paper, we investigate how navigation and collision avoidance mechanisms can instead be learned from scratch, via a continuous, self-supervised learning process using a simple monocular camera.

Learning offers considerable promise for mobile robotic systems: by observing the outcomes of navigational decisions in the real world, mobile robots can continuously improve their proficiency and adapt to the statistics of natural environments. Not only can learning-based systems lift some of the assumptions of geometric reconstruction methods, but they offer two major advantages that are not present in analytic approaches: (1) learning-based methods adapt to the statistics of the environments in which they are trained and (2) learning-based systems can learn from their mistakes. The first advantage means that a learning-based navigation system may be able to act more intelligently even under partial observation by exploiting its knowledge of statistical patterns. The second advantage means that, when a learning-based system does make a mistake that results in a failure, the resulting data can be used to improve the system to prevent such a failure from occurring in the future. This second advantage, which is the principal focus of this work, is closely associated with reinforcement learning: algorithms that learn from trial-and-error experience.

\begin{wrapfigure}{r}{0.5\columnwidth}
  \vspace*{-10pt}
  \centering
  \includegraphics[width=0.8\columnwidth,angle=-90]{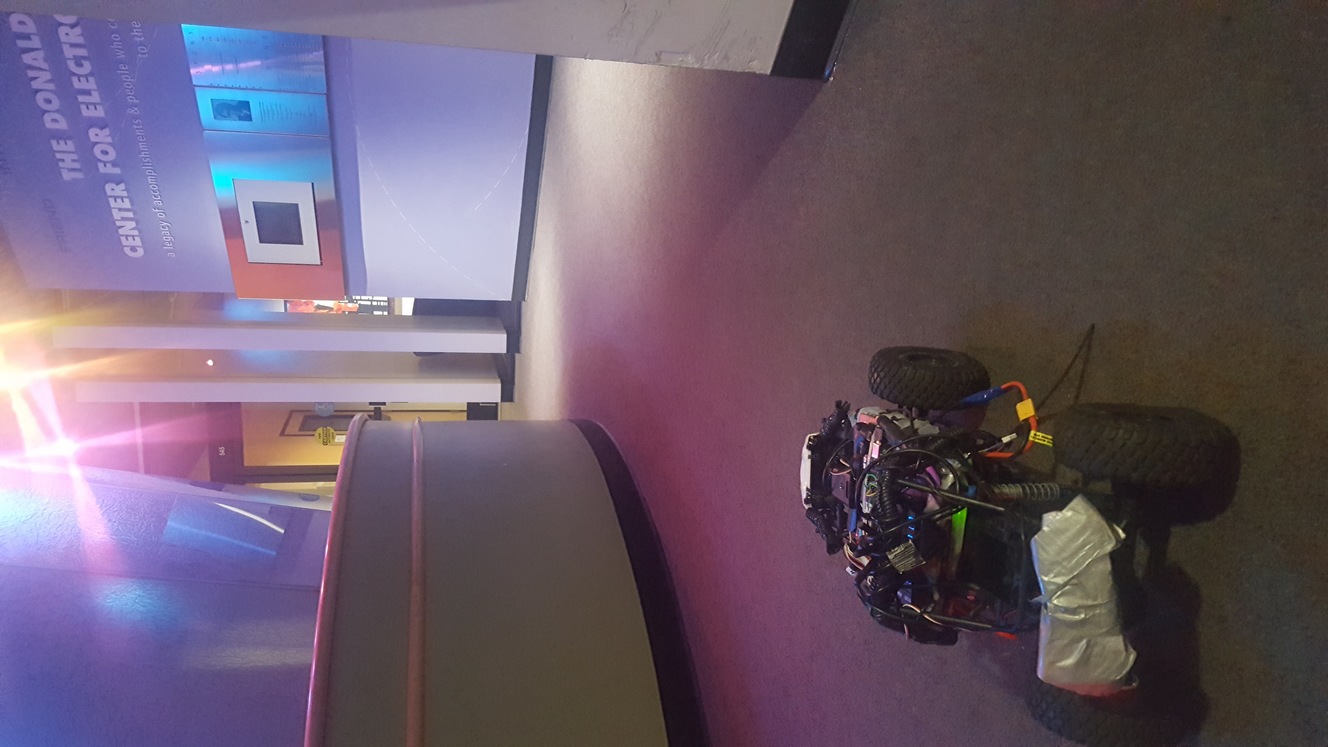}
  \caption{Our RC car navigating Cory Hall from raw monocular camera images using our learned navigation policy trained using 4 hours of fully autonomous reinforcement learning.}
  \vspace*{-10pt}
  \label{fig:teaser}
\end{wrapfigure}

Reinforcement learning methods are typically classified as either model-free or model-based. Value-based model-free approaches learn a function that takes as input a state and action, and outputs the value (i.e., the expected sum of future rewards). Policy extraction is then performed by selecting the action that maximizes the value function. Model-based approaches learn a predictive function that takes as input a state and a sequence of actions, and output future states. Policy extraction is then performed by selecting the action sequence that maximizes the future rewards using the predicted future states. In general, model-free algorithms can learn complex tasks~\cite{Mnih2013_NIPS,Lillicrap2016_ICLR}, but are usually sample-inefficient, while model-based algorithms are typically sample-efficient~\cite{Deisenroth2011_ICML}, but have difficulty scaling to complex, high-dimensional tasks~\cite{deisenroth2012toward}.

We explore the intersection between value-based model-free algorithms and model-based algorithms in the context of learning robot navigation policies. Our work has three primary contributions. The first contribution is a generalized computation graph for reinforcement learning that subsumes value-based model-free methods and model-based methods. The second contribution is instantiations of the generalized computation graph for the task of robot navigation, resulting in a suite of hybrid model-free, model-based algorithms. The third contribution is an extensive empirical evaluation in which we: (a) investigate and discover design decisions regarding our robot navigation computation graph that are important for stable and high performing policies; (b) show our approach learns better policies than state-of-the-art reinforcement learning methods in a simulated environment; and (c) show our approach can learn a successful navigation policy on an RC car in a real-world, indoor environment from scratch using raw monocular images (Fig.~\ref{fig:teaser}).


\section{Related Work}
\label{sec:related}

There is extensive prior work on autonomous robot navigation, ranging from indoor mobile robots~\cite{rosen1968application} to full-sized vehicles~\cite{thorpe1988vision,urmson2008autonomous}. Many of these prior works operate under the paradigm of maintaining a map while simultaneously localizing and planning actions. While simultaneous localization and mapping (SLAM)~\cite{ThrBurFox2008} and planning approaches have had difficulties with the size, weight, and power (SWaP) constraints of mobile robots~\cite{Leonard1991_IROS}, recent work has demonstrated these algorithms can be run on SWaP constrained platforms, such as quadrotors~\cite{shen2011autonomous}. However, methods based on explicit modeling of the environment still have a number of limitations, including difficulties with dynamic environments, textureless scenes, and high-bandwidth sensors~\cite{fuentes2015visual}.

Learning-based methods have attempted to address these limitations by learning from data. These supervised learning methods include learning: drivable routes and then using a planner~\cite{barnes2017find}, near-to-far obstacle detectors~\cite{hadsell2009learning}, reactive controllers on top of a map-based planner~\cite{Richter2017_RSS}, driving affordances~\cite{chen2015deepdriving}, and end-to-end driving from demonstrations~\cite{pomerleau1989alvinn,bojarski2016end}. However, the capabilities of powerful and expressive models like deep neural networks are often constrained in large part by the available data, and methods based on human supervision are inherently limited by the amount of human data available. In constrast, our work takes a self-supervised approach, learning directly from real-world experience. In principle, a fully autonomous system of this sort can improve continuously, collect more data over its entire lifetime, and correct its own mistakes over time.

Autonomously learning from trial-and-error is the hallmark of reinforcement learning. Model-free reinforcement learning methods have been able to learn complex tasks~\cite{Mnih2013_NIPS,Lillicrap2016_ICLR}, but are typically less sample-efficient, while model-based methods are usually more sample-efficient, but have difficulty with high-bandwidth sensors, such as cameras, and complex environments~\cite{deisenroth2012toward}. While these methods have been used to learn robot navigation policies, they often require simulation experience~\cite{michels2005high,sadeghi2017cad}. In contrast, our approach learns from scratch to navigate using monocular images solely in the real-world.

Our generalized computation graph allows for model instantiations that combine model-free and model-based approaches. Combining model-free and model-based methods has been investigated in a number of prior works~\cite{Sutton1991_AAAI}. Prior work has explored value function estimators that take in multiple actions~\cite{mujika2016multi,oh2017value}, in the context of simulated tasks such as playing Atari games. In contrast to this prior work, we examine a variety of multi-action prediction models, trained both with supervised learning and Q-learning style methods, and demonstrate effective learning in complex real-world environments. Our empirical results show that the design presented in prior work~\cite{mujika2016multi,oh2017value} is often not the best one for real-world continuous learning tasks, and shed light on the tradeoffs involved with single- and multi- action Q-learning, as well as purely prediction-based control~\footnote{We note that~\cite{Kahn2017_arxiv} uses a purely prediction-based method, but focuses on uncertainty and safety.}.


\section{Preliminaries}
\label{sec:pre}

Our goal is to learn collision avoidance policies for mobile robots. We formalize this task as a reinforcement learning problem, where the robot is rewarded for collision-free navigation.

In reinforcement learning, the goal is to learn a policy that chooses actions $\ba_t \in \mathcal{A}$ at each time step $t$ in response to the current state $\bs_t \in \mathcal{S}$, such that the total expected sum of discounted rewards is maximized over all time. At each time step, the system transitions from $\bs_t$ to $\bs_{t+1}$ in response to the chosen action $\ba_t$ and the transition probability $T(\bs_{t+1} | \bs_t, \ba_t)$, collecting a reward $r_t$ according to the reward function $R(\bs_t, \ba_t)$. The expected sum of discounted rewards is then defined as $\mathrm{E}_{\pi,T}[\sum_{t'=t}^\infty \gamma^{t'-t} r_{t'} | \bs_t, \ba_t]$, where ${\gamma \in [0,1]}$ is a discount factor that prioritizes near-term rewards over distant rewards, and the expectation is taken under the transition function $T$ and a policy $\pi$. Algorithms that solve the reinforcement learning problem are typically either model-free or model-based. The generalized computation graph we introduce subsumes value-based model-free methods and model-based, therefore we will first provide a brief overview of both these methods.


\subsection{Value-based model-free reinforcement learning}
\label{sec:pre-mf}

Value-based model-free algorithms learn a value function in order to select which actions to take. In this work, we will focus specifically on algorithms that learn state-action value functions, also called Q-functions. The standard parametric Q-function, $Q_\theta(\bs, \ba)$, is a function of the current state and a single action, and outputs the expected discounted sum of future rewards that will be received by the optimal policy after taking action $\ba$ in state $\bs$, where $\theta$ denotes the function parameters. If the Q-function can be learned, then the optimal policy can be recovered simply by taking that action $\ba$ that maximizes $V_t = Q_\theta(\bs_t, \ba)$. A standard method for learning the Q-function is to minimize the Bellman error, given by

{\small
\vspace*{-10pt}
\begin{align*}
\mathcal{E}_t(\theta) = \frac{1}{2}\mathrm{E}_{\bs \sim T, \ba \sim \pi}\left[ \| r_t + \gamma V_{t+1} - Q_\theta(\bs_t,\ba_t) \|^2 \right], 
\end{align*}
}%
where the actions are sampled from $\pi(\cdot | \bs)$ and the $V_{t+1}$ term is known as the bootstrap. The policy $\pi$ can in principle be any policy, making Q-learning an off-policy algorithm.

Multi-step returns~\cite{SuttonBarto} can be incorporated into Q-learning and other TD-style algorithms by summing over the rewards over $N$ steps, and then using the current or target Q-function to label the $N$+1th step. Multi-step returns can increase sample efficiency, but also make the algorithm on-policy. Defining the $N$-step value as $V^{(N)}_t = \sum_{n=0}^{N-1} \gamma^{n} r_{t+n} + \gamma^N V_{t+N}$, we can augment the standard Bellman error minimization objective by considering a weighted combination of Bellman errors from horizon length 1 to $N$:
\begin{align*}
\mathcal{E}_t(\theta) = \frac{1}{2}\mathrm{E}_{\bs \sim T, \ba \sim \pi}&\left[ \| \sum_{N'=1}^{N} w_{N'} V^{(N')}_t - Q_\theta(\bs_t,\ba_t) \|^2 \right] \;: \\
&\sum_{N'=1}^{N}w_{N'} = 1, w_{N'} \geq 0. 
\end{align*}%


\subsection{Model-based reinforcement learning}
\label{sec:pre-mb}

In contrast to model-free approaches, which avoid modelling the transition dynamics by learning a Q-value and using bootstrapping, model-based approaches explicitly learn a transition dynamics $\hat{T}_\theta(\bs_{t+1} | \bs_t, \ba_t)$ parameterized by vector $\theta$. At time $t$ in state $\bs_t$, the next action $\ba_t$ is selected by solving the finite-horizon control problem

{\small
\vspace*{-10pt}
\begin{align*}
\arg\max_{\bA_t^H} \mathrm{E} &\left[ \sum_{h=0}^{H-1} \gamma^h R(\hat{\bs}_{t+h}, \ba_{t+h}) \right] : \nonumber \\
&\hat{\bs}_{t'+1} \sim \hat{T}_\theta(\hat{\bs}_{t'+1} | \hat{\bs}_{t'}, \ba_{t'}), \hat{\bs}_t = \bs_t,
\end{align*}
}%
in which $H$ is the planning horizon and ${\bA_t^H = (\ba_t, \ba_{t+1}, ..., \ba_{t+H-1})}$ is the planned action sequence. Note that the reward function $R$ must be known a priori.

Since planning for large $H$ is expensive and often undesirable due to model inaccuracies, planning is typically done in a model predictive control (MPC) fashion in which the optimization is solved at each time step, the first action from the optimized action sequence is executed, and the process is repeated. Standard model-based algorithms then alternate between gathering samples using MPC and storing transitions $(\bs_t, \ba_t, \bs_{t+1})$ into a dataset $\mathcal{D}$, and updating the transition model parameter vector $\theta$ to maximize the likelihood of the transitions stored in $\mathcal{D}$.


\subsection{Comparing model-free and model-based methods}
\label{sec:pre-compare}

We now compare the advantages and disadvantages of both model-free and model-based methods for learning continuous robot navigation policies by evaluating three metrics: sample efficiency, stability, and final performance.

Model-free methods have empirically demonstrated state-of-the-art performance in many complex tasks~\cite{Mnih2013_NIPS,Lillicrap2016_ICLR}, including navigation~\cite{gupta2017cognitive}. However, model-free techniques are often sample inefficient. Specifically, for ($N$-step) Q-learning, bias from bootstrapping and high variance multi-step returns can lead to slow convergence~\cite{Thrun1993_CMSS}. Furthermore, Q-learning often requires experience replay buffers and target networks for stable learning, which also further decreases sample efficiency~\cite{Mnih2013_NIPS}. We empirically demonstrate that these stability issues are further exacerbated in continuous learning scenarios, such as robot navigation.

In contrast, model-based methods can be very sample efficient and stable, since learning the transition model reduces to supervised learning of dense time-series data~\cite{Deisenroth2011_ICML}. However, final performance can be poor because maximizing the accuracy of the learned transition model is merely a surrogate objective, that is to say that an accurate transition model does not necessarily mean the policy will perform well. In addition, all three metrics suffer when the state space is high-dimensional, such as when learning from raw images~\cite{oh2015action}.

In order to develop a sample efficient, stable, and high performing reinforcement learning algorithm for training robot navigation policies, we will leverage aspects of both model-free and model-based methods.


\section{A generalized computation graph for reinforcement learning}
\label{sec:gcg}

We will now introduce a generalized computation graph for reinforcement learning that subsumes both model-free value function-based methods and model-based algorithms. This generalized computation graph not only encompasses existing model-free and model-based methods, but also will allow us to formulate a sample-efficient, stable, and high-performing algorithm for training robot navigation policies.

\begin{figure}[b]
\centering
\vspace*{-15pt}
\includegraphics[width=0.95\columnwidth]{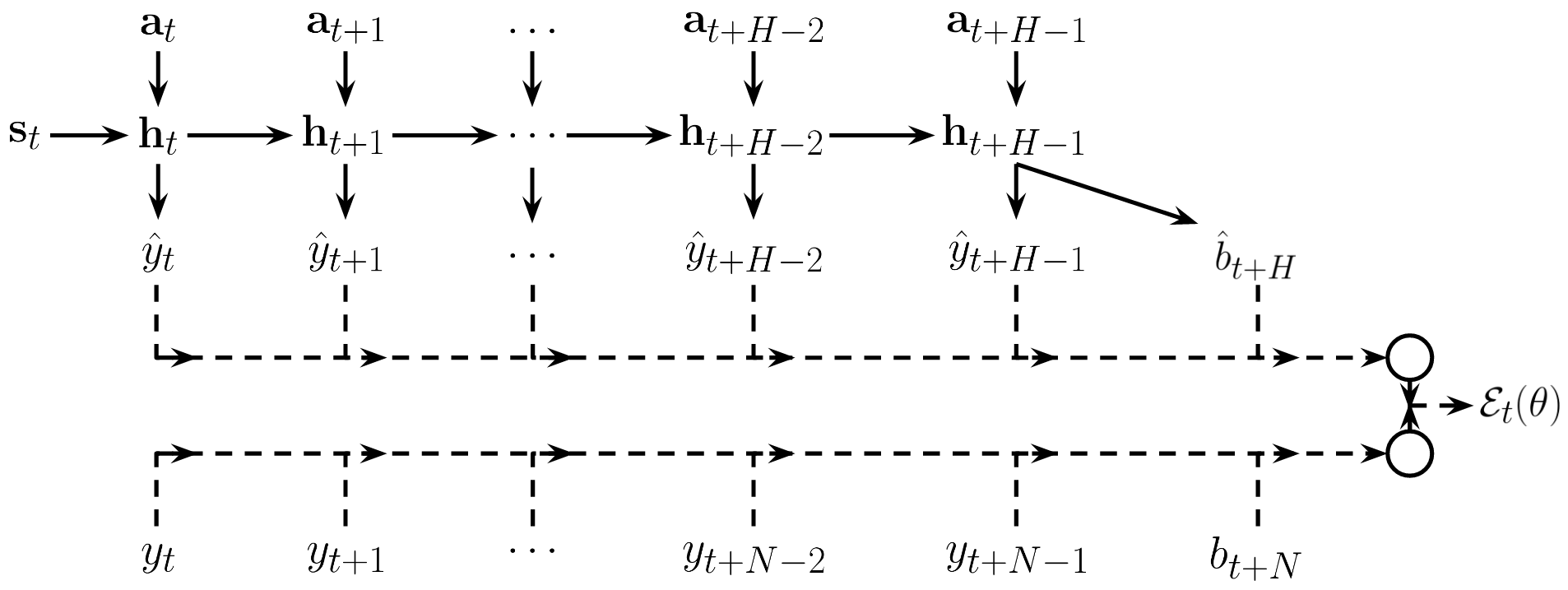}
\caption{A generalized computation graph for model-free, model-based, and hybrid reinforcement learning algorithms. These algorithms train models that take as input the current state and a sequence of $H$ actions and produce $H$ outputs plus a final terminal output. These predicted outputs are then combined and compared with $N$ labels to produce an error signal that is used by an optimizer in order to train the model. Solid lines indicate computations involving model parameters $\theta$ while dashed lines indicate signal flow.}
\label{fig:gen-comp-graph}
\end{figure}

Fig.~\ref{fig:gen-comp-graph} shows a generalized computation graph for reinforcement learning models. The computation graph $G_\theta(\bs_t, \bA_t^H)$ parameterized by vector $\theta$ takes as input the current state $\bs_t$ and a sequence of $H$ actions ${\bA_t^H = (\ba_t, ..., \ba_{t+H-1})}$ and produces $H$ sequential predicted outputs ${\hat{Y}_t^H = (\hat{y}_t, ..., \hat{y}_{t+H-1})}$ and a predicted terminal output $\hat{b}_{t+H}$. These predicted outputs $\hat{Y}_t^H$ and $\hat{b}_{t+H}$ are combined and compared with labels $Y_t^N$ and $b_{t+N}$ to form an error signal $\mathcal{E}_t(\theta)$ that is minimized using an optimizer.

We will now show how the generalized computation graph can be instantiated to be standard model-free value function-based methods and model-based methods. We first instantiate the computation graph for $N$-step Q-learning by letting $y$ be reward and $b$ be the future value estimate; setting the model horizon $H=1$ and using $N$-step returns; and letting the error function be the Bellman error: ${\mathcal{E}_t(\theta) = \| (\hat{y}_t + \gamma \hat{b}_{t+1}) - (\sum_{n=0}^{N-1} \gamma^n y_{t+n} + \gamma^N b_{t+N}) \|_2^2}$. Next, we instantiate the computation graph for standard model-based learning by ignoring $y$ and letting $b$ be the state; setting the model horizon $H=1$ and label horizon $H=1$; and letting the error function minimize the difference between the predicted and actual next state: ${\mathcal{E}_t(\theta) = \| \hat{b}_{t+1} - b_{t+1} \|_2^2}$.

In order to use the generalized computation graph in a reinforcement learning algorithm, we must be able to extract a policy from the generalized computation graph. We define $J(\bs_t, \bA_t^H)$ to be the generalized policy evaluation function, which is a scalar function such that $\pi(\bA^H | \bs_t) = \arg\max_{\bA^H} J(\bs_t, \bA^H)$. Similar to before, we now instantiate $J$ for standard model-free value function-based methods and model-based methods. For $N$-step Q-learning, $J(\bs_t, \ba_t) = \hat{y}_t + \gamma \hat{b}_{t+1}$ is the estimated future value. For standard model-based learning, $J(\bs_t, \ba_t) = R(\bs_t, \ba_t) + J(\hat{b}_{t+1}, \arg\max_{\ba} J(\hat{b}_{t+1}, \ba))$ is the reward function evaluated on the single-step dynamics model propagated from the current state for multiple timesteps.

Using the generalized computation graph $G_\theta$, the graph error function $\mathcal{E}$, and the policy evaluation function $J$, we outline a general reinforcement learning algorithm in Alg.~\ref{alg:generalized-rl}. This framework for general-purpose predictive learning can subsume both standard model-free value function-based methods and model-based algorithms.

\begin{algorithm}[t]
\caption{\small{Reinforcement learning with generalized computation graphs}}
\label{alg:generalized-rl}
\begin{algorithmic}[1]
\STATE \textbf{input}: computation graph $G_\theta(\bs_t, \bA_t^H)$, error function $\mathcal{E}_t(\theta)$, and policy evaluation function $J(\bs_t, \bA_t^H)$
\STATE initialize dataset $\mathcal{D} \leftarrow \emptyset$
\FOR{$t=1$ to $T$}
	\STATE get current state $\bs_t$
	\STATE $\bA_t^H \leftarrow \arg\max_{\bA} J(\bs_t, \bA)$
	\STATE execute first action $\ba_t$
	\STATE receive labels $y_t$ and $b_t$
	\STATE add $(\bs_t, \ba_t, y_t, b_t)$ to dataset $\mathcal{D}$
	\STATE update $G_\theta$ by ${\theta \leftarrow \arg\min_\theta \mathcal{E}_{t'}(\theta)}$ using $\mathcal{D}$
\ENDFOR
\end{algorithmic}
\end{algorithm}


\section{Learning navigation policies with self-supervision}
\label{sec:nav}

The computation graph outlined in the previous section can be instantiated to perform both fully model-free learning, where the model predicts the expected sum of future rewards, and fully model-based learning, where the model predicts future states. However, in practical robotic applications, especially in robotic mobility, we often have prior knowledge about our system. For example, the dynamics of a car could be identified in advance with considerable accuracy. Other aspects, such as the relationship between observed images and positions of obstacles, are exceptionally difficult to specify analytically, and could be learned by a model. The question then arises: which aspects of the system should we learn to predict, which aspects should we handle with analytic models, and which aspects should we ignore?

We will now explore the space of possible instantiations of the generalized computation graph for learning robot navigation policies. While some design decisions will remain constant, other design choices will have multiple options, which we will describe in detail and empirically evaluate in our experiments.

\begin{figure*}
\centering
\includegraphics[width=0.95\textwidth]{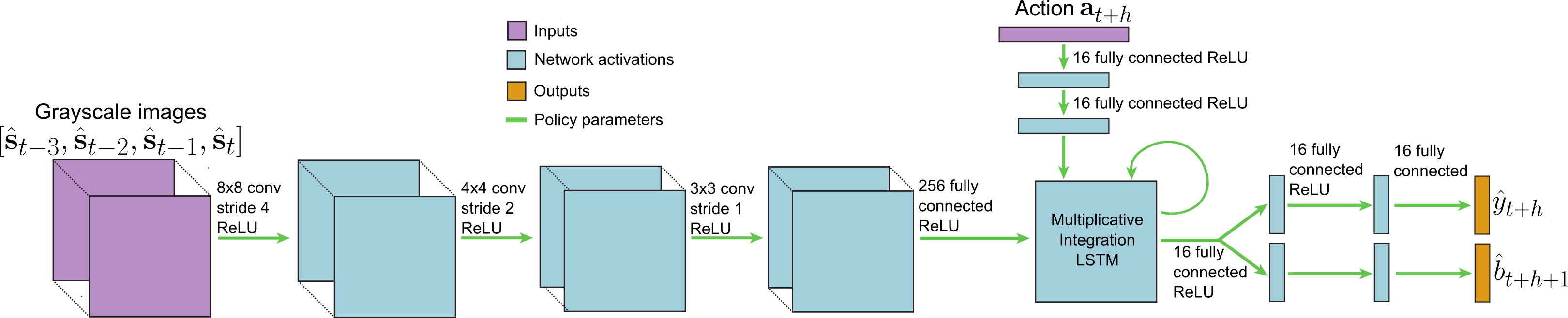}
\caption{Recurrent neural network computation graph for robot navigation policies. The network takes as input the past four grayscale images, which are processed by convolutional layers to become the initial hidden state for the recurrent unit. The recurrent unit is a multiplicative integration LSTM~\cite{wu2016multiplicative}. From $h=0$ to $H-1$, the recurrent unit takes as input the processed action and the previous hidden state, and produces the next hidden state and the RNN output vector. The RNN output vector is passed through final layers to produce the model outputs $\hat{y}_{t+h}$ and $\hat{b}_{t+h+1}$.}
\label{fig:nav-graph}
\vspace*{-15pt}
\end{figure*}

\textbf{Model parameterization.} While many function approximators could be used to instantiate our generalized computation graph, the function approximator needs to be able to cope with high-dimensional state inputs, such as images, and accurately model sequential data due to the nature of robot navigation. We therefore parameterize the computation graph as a deep recurrent neural network (RNN), depicted in Fig.~\ref{fig:nav-graph}. 

\textbf{Model outputs.} While we have defined what our deep recurrent neural network model takes as input, namely the current state and a sequence of actions, we need to specify what quantities are the model outputs $\hat{y}$ and $\hat{b}$. We consider two quantities. The first quantity is the standard approach in the reinforcement literature: $\hat{Y}_t^H$ represent rewards and $\hat{b}_{t+H}$ represents the future value-to-go. For the task of collision-free navigation, we define the reward as the robot's speed, which is typically known using onboard sensors, and therefore the value is approximately the distance the robot travels before experiencing a collision. The advantage of outputting the value-to-go is that this is precisely what our agents want to maximize. However, the value representation does not leverage any prior knowledge about robot navigation.

The second quantity we propose is specific to collision avoidance, in which $\hat{Y}_t^H$ represents the probability of collision at or before each timestep---that is, $\hat{y}_{t+h}$ is the probability the robot will collide between time $t$ and $t+h$---and $\hat{b}_{t+H}$ represents the best-case future likelihood of collision. One advantage of outputting collision probabilities is that this binary signal may be easier and faster to learn.

\textbf{Policy evaluation function.} Given the outputs of the navigation computation graph, we now need to define how the task of collision-free robot navigation is encoded into the policy evaluation function $J$.

If the model output quantities are values, which in our case is the expected distance-to-travel, then the policy evaluation function is simply the value ${J(\bs_t, \bA_t^H) = \sum_{h=0}^{H-1} \gamma^h \hat{y}_{t+h} + \gamma^H \hat{b}_{t+H}}$.

If the model output quantities are collision probabilities, then the policy evaluation function needs to somehow encourage the robot to move through the environment. We assume that the robot will be travelling at some fixed speed, and therefore the policy evaluation function needs to evaluate which actions are least likely to result in collisions ${J(\bs_t, \bA_t^H) = \sum_{h=0}^{H-1} - \hat{y}_{t+h} - \hat{b}_{t+H}}$.

\textbf{Policy evaluation.} Using the policy evaluation function, action selection is performed by solving the finite-horizon planning problem $\arg\max_{\bA^H} J(\bs_t, \bA^H)$. Although we can use any optimal control or planning algorithm to perform the maximization, in our experiments we use a simple random shooting method, in which the $K$ randomly sampled action sequences are evaluated with $J$ and the action sequence with the largest value is chosen. We also evaluated action selection using the cross entropy method~\cite{deng2006cross}, but empirically found no difference in performance. However, exploring other methods could further improve performance.

\textbf{Model horizon.} An important design decision is the model horizon $H$. The value of the model horizon in effect determines the degree to which the model is model-free or model-based. For $H=1$, the model is fully model-free because it does not model the dynamics of the output, while for horizon $H$ that is the full length of a (possibly infinite) episode, the model is fully model-based in the sense that the model has learned the dynamics of the output. For intermediate values of $H$, the model is a hybrid of model-free and model-based methods. We empirically evaluate different horizon values in our experiments.

\textbf{Label horizon.} In addition to the model horizon, we must decide the label horizon $N$. The label horizon $N$ can either be set to the model horizon $H$, or to some value $N>H$. Although setting the label horizon $N$ to be larger than the model horizon $H$ can increase learning speed, as $N$-step Q-learning often demonstrates, the learning algorithm then becomes on-policy. This is an undesirable property for robot navigation because we would like our policy to be able to be trained with any kind of data, including data gathered by old policies or by exploration policies. We therefore set the label horizon $N$ to be the same as the model horizon $H$.

\textbf{Bootstrapping.} Because we chose the label horizon to be the same as the model horizon, the only way in which the model can learn about future outcomes is by increasing the model horizon or by using bootstrapping. An advantage of increasing the model horizon is that the model becomes more model-based, which has been shown to be sample efficient. However, increasing the model horizon increases the difficulty of policy evaluation because the search space grows exponentially in $H$. Bootstrapping can alleviate the planning problem by allowing for smaller $H$, but bootstrapping can cause bias and instability in the learning process. We evaluate the effect of bootstrapping in our experiments.

\textbf{Training the model.} Finally, to train our model using a dataset $\mathcal{D}$, we need to define the loss function between the model outputs and the labels.
Using samples $(\bs_t^H, \bA_t^H, y_t^H) \in \mathcal{D}$ from the dataset, if the model outputs and labels are values, the loss function is the standard Bellman error {\small${\mathcal{E}_t(\theta) = \| \sum_{n=0}^{N-1} \gamma^n y_{t+n} + \gamma^N b_{t+H} - J(\bs_t, \bA_t^H) \|_2^2}$}, in which $b_{t+H} = \max_{\bA^H} J(\bs_{t+H}, \bA^H)$.
If the model outputs are collision probabilities, the loss function is the cross entropy loss

{\small
\vspace*{-20pt}
\begin{align*}
\mathcal{E}_t(\theta) = -\big[\sum_{h=0}^{H-1} &y_{t+h} \log(\hat{y}_{t+h}) + (1 - y_{t+h}) \log(1 - \hat{y}_{t+h}) + \\
&b_{t+H} \log(\hat{b}_{t+H}) + (1 - b_{t+H}) \log(1 - \hat{b}_{t+H})\big],
\end{align*}
}%
in which $b_{t+H} = \min_{\bA^H} \frac{1}{H} \sum_{h=0}^H \hat{y}_{t+H+h}$ represents the lowest average probability of collision the robot can achieve at time $t+H$. We note that these probabilities can also be learned using a mean squared error loss, and we examine the effect of the loss function choice in our experiments.


\section{Experiments}
\label{sec:exp}

The navigation computation graphs discussed in the previous section can be instantiated in various ways, such as predicting full reward or only collision, and predicting with different horizons. In our experiments, we aim to evaluate the various design choices exposed by the generalized computation graph framework, including the special cases that correspond to standard algorithms such as Q-learning, and study their impact on both simulated and real-world robotic navigation. In our evaluations, we aim to answer the following questions:

\newcommand{\Qdesign}{\textbf{Q1}}
\newcommand{\Qcompare}{\textbf{Q2}}
\newcommand{\Qrealworld}{\textbf{Q3}}

\begin{enumerate}
\item[\Qdesign] How do the different design choices for our navigation computation graph affect performance?
\item[\Qcompare] Given the best design choices, how does our approach compare to prior methods?
\item[\Qrealworld] Is our approach able to successfully learn a navigation policy on a real robot in a complex environment?
\end{enumerate}

Experiment videos and code are provided on our website \url{github.com/gkahn13/gcg}.


\subsection{Simulation results}
\label{sec:exp-sim}

\begin{figure}[b]
  \vspace*{-10pt}
  \centering
  \includegraphics[width=0.48\columnwidth,trim={1cm 2cm 3cm 2cm},clip]{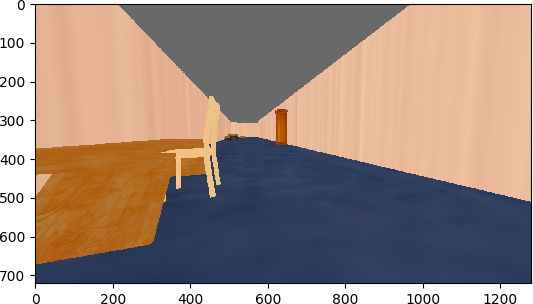}
  \includegraphics[width=0.48\columnwidth,trim={1cm 2cm 3cm 2cm},clip]{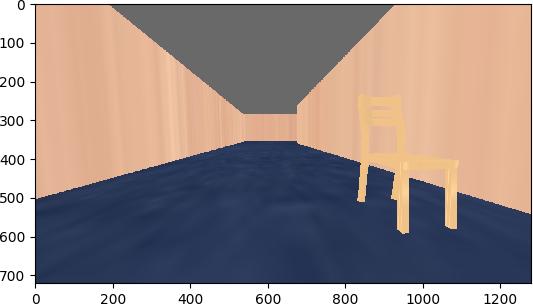}
  \caption{First-person view images from a simulated RC car learning to navigate in a cluttered hallway.}
  \vspace*{0pt}
  \label{fig:cluttered-hallway}
\end{figure}

We first present results on a simulated RC car in a cluttered, indoor environment (Fig.~\ref{fig:cluttered-hallway}). The RC car was created using the Bullet physics simulator and images were rendered using the Panda3d graphics engine~\cite{goslin2004panda3d}. The robot state $\mathcal{S} \in \mathbb{R}^{2304}$ is a $64 \times 36$ grayscale image taken from an onboard forward-facing camera. The car navigates at a fixed speed of 2m/s, therefore the action space $\mathcal{A} \in \mathbb{R}^1$ consists solely of the steering angle. The car observes the current image and selects an action every $\textnormal{dt}=0.25$ seconds. We note that 1 hour of simulator time will result in $14,400$ datapoints. We define an episode as the car acting in the environment until it either crashes or travels 1000 meters. Because we are considering the setting of continuous learning, each episode continues from where the previous episode ended; if the previous episode ended in a collision, the car first executes a hard-coded backup maneuver before starting the next episode. All experiments were evaluated three times with different random seeds.

\textbf{Evaluating design decisions for robot navigation learning.}
We will now explore and empirically evaluate the four design decisions (Sec.~\ref{sec:nav}) of our navigation computation graph: the model output, loss function, model horizon, and bootstrapping. These decisions will be evaluated in terms of their effect on sample efficiency, stability, and final performance.

\subsubsection{Model outputs and loss function}
\begin{wrapfigure}{r}{0.6\columnwidth}
  \vspace*{-10pt}
  \centering
  \includegraphics[width=0.6\columnwidth]{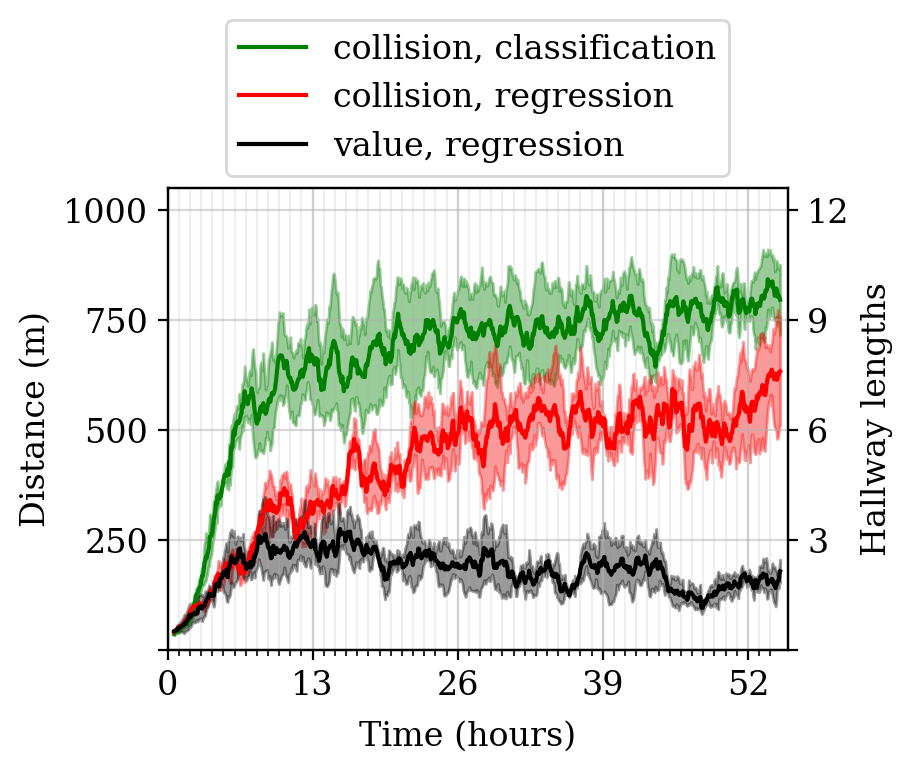}
  \caption{Evaluation of our approach with different model outputs (values or collision probabilities) and training methods (regression or classification).}
  \label{fig:sim-dd-outputs-loss}
  \vspace*{-10pt}
\end{wrapfigure}

Fig.~\ref{fig:sim-dd-outputs-loss} shows learning curves for different model outputs and loss functions. ``Value'' corresponds to outputs that represent the expected sum of future rewards, while ``collision'' corresponds to outputs that represent probabilities of collision. Regression corresponds to using a mean squared error loss function, while classification corresponds to using a cross entropy loss function. For consistency, these models all use a long horizon ($H=16$) and do not use bootstrapping.

These results show that, given the same mean squared error loss function, outputting collision probabilities leads to substantially more sample-efficient and higher-performing policies than outputting predicted values. Although one might be tempted to say that the improved performance is due to different policy evaluation functions $J$, we note that, because the car is always moving at a fixed speed of 2m/s and the reward function is the car's speed, the only rewards the value model trains on are either 2 (if no collision) or 0 (if collision), which is a binary signal. The major difference lies in the form of the mean squared error loss function: the value model loss is a single loss on the sum of the outputs, while the collision model is the sum of $H$ separate losses on each of the outputs. The collision model therefore has additional supervision about when the collision labels occur in time. Additionally, training with a cross entropy loss is significantly better than training with a mean squared error loss in terms of both sample efficiency and final performance.

This comparison shows that predicting discrete future events can lead to faster and more stable learning than predicting continuous sums of discounted rewards. While we have only shown this finding in the context of robot navigation, this insight could lead to a new class of sample-efficient, stable, and high-performing reinforcement learning algorithms~\cite{Bellemare2017_ICML}.

\begin{figure}[t]
\centering
\includegraphics[width=0.95\columnwidth]{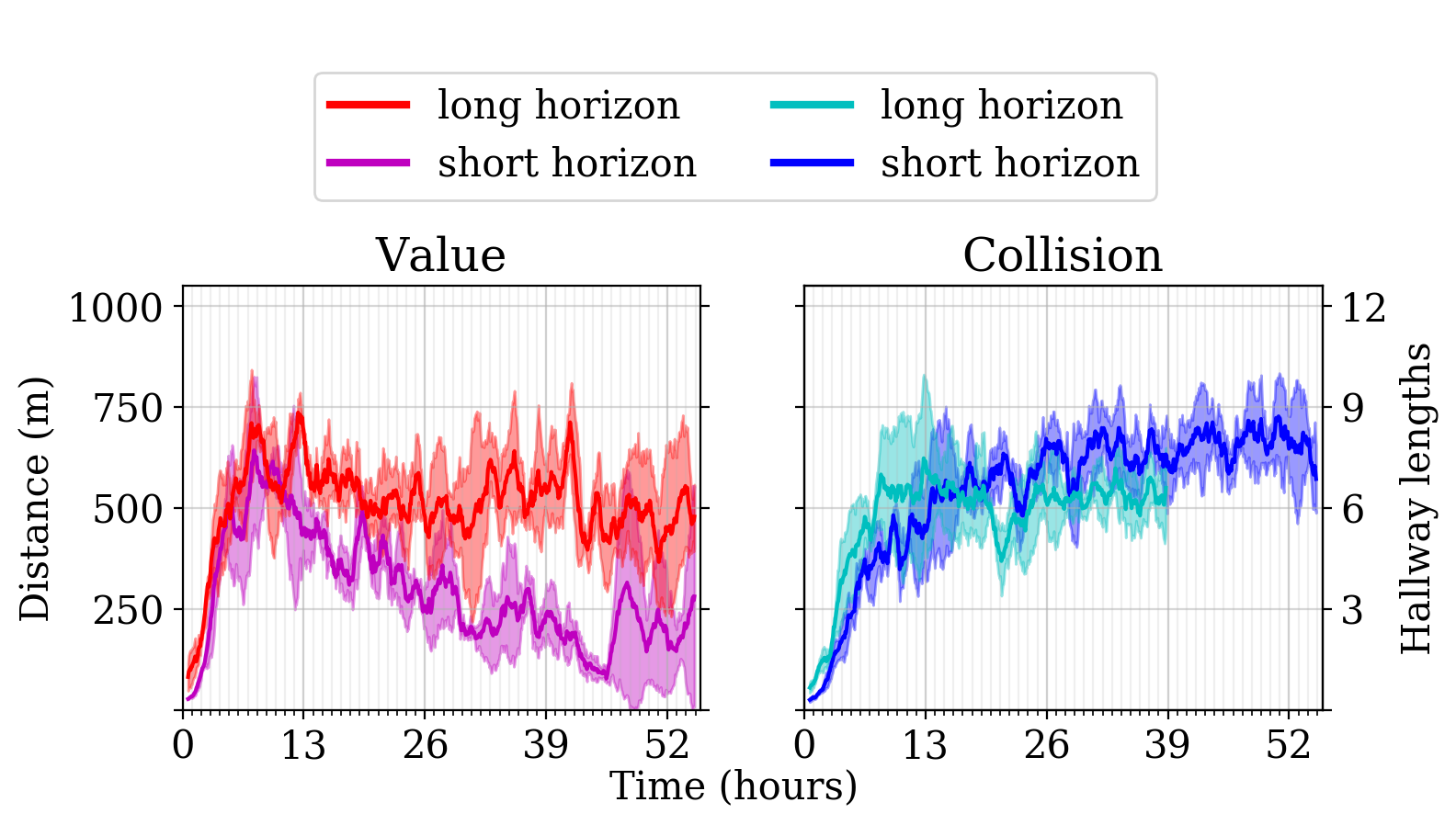}
\caption{Evaluation of our approach with different model horizons. (Collision with long horizon was terminated early due to computational constraints.)}
\label{fig:sim-dd-horizon}
\vspace*{-20pt}
\end{figure}

\begin{figure}[b]
\vspace*{-20pt}
\centering
\includegraphics[width=0.95\columnwidth]{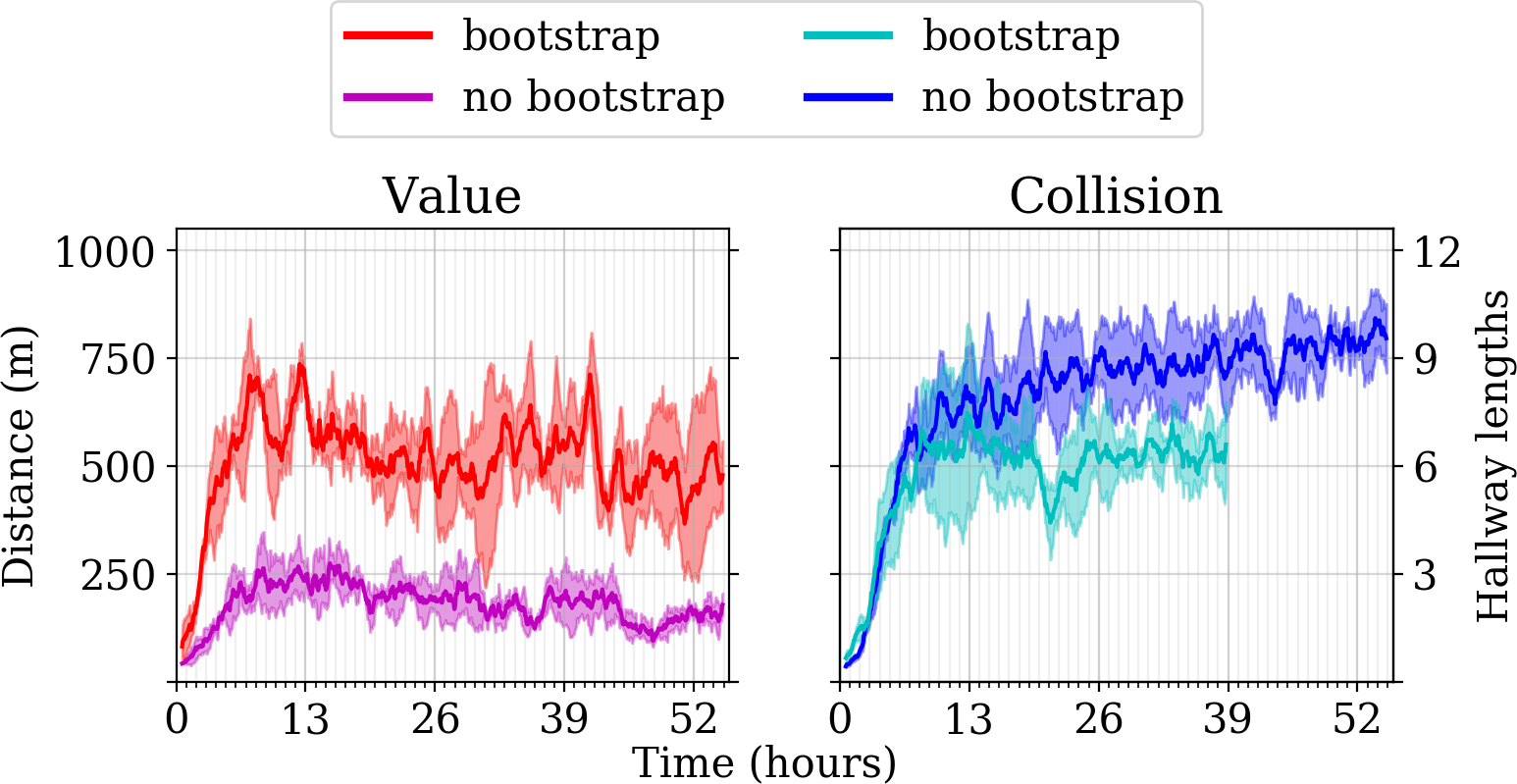}
\caption{Evaluation of our approach with and without bootstrapping.}
\label{fig:sim-dd-bootstrapping}
\end{figure}

\begin{figure*}[t]
\centering
\includegraphics[width=0.92\textwidth]{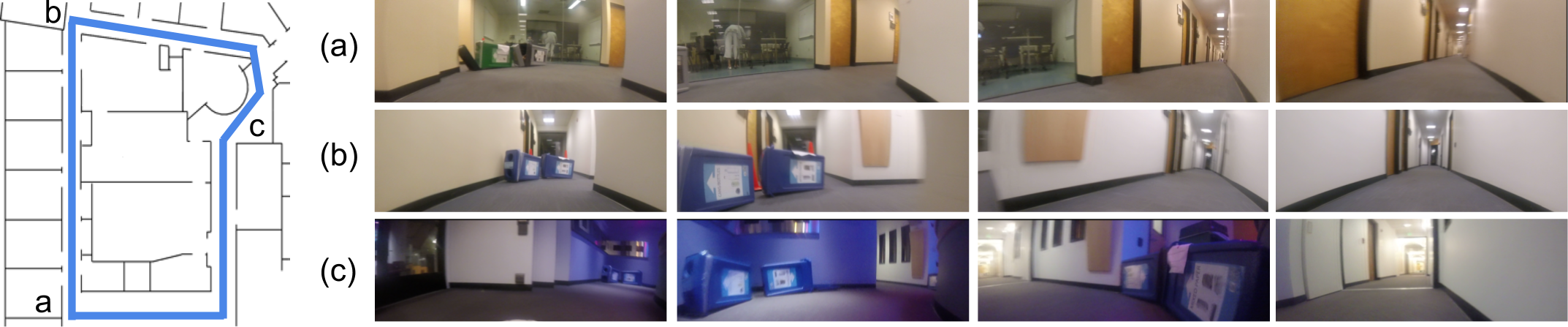}
\caption{Real-world RC car experiments on the 5th floor of Cory hall. The path the robot can follow is drawn on the floor plan (left), however, this path is not provided to the RC car. Three example trajectories of the RC car navigating with our learned policy are shown.}
\label{fig:rw-pics}
\vspace*{-20pt}
\end{figure*}

\subsubsection{Model horizon}
We next examine the effect of using short model horizons ($H=1$, corresponding to a $\frac{1}{2}$m lookahead) versus long model horizons ($H=16$, corresponding to an 8m lookahead). For consistency, and so that the model with the short horizon can learn about events beyond its planning horizon, all models use bootstrapping. Fig.~\ref{fig:sim-dd-horizon} shows these results for models that output values and models that output collision probabilities. The models use regression for outputs that are values and classification for outputs that are collision probabilities.

For models that output values, training with a longer horizon is more stable and leads to a higher performing final policy. While the short horizon model initially has the same learning speed, its performance peaks early on and declines thereafter. While one might be inclined to attribute this decrease in performance to overfitting, we note that the long horizon model should be even more prone to overfitting, yet it performs much better. We therefore conclude the longer horizon model learns better because the long horizon mitigates the bias of the bootstrap due to the exponential weighting factor $\gamma^H$ in front of the bootstrap term.

However, for models that output collision probabilities, we do not notice any change in performance when comparing short and long horizon models. This could be due to the fact that the probabilities are necessarily bounded between 0 and 1, which minimizes the bias from bootstrapping. Future work investigating the relationship between classification and bootstrapping could yield more stable and sample-efficient reinforcement learning algorithms.

\subsubsection{Bootstrapping}
Finally, we investigate the effect of bootstrapping. Fig.~\ref{fig:sim-dd-bootstrapping} shows these results for models that output values and models that output collision probabilities. The models all use long horizon prediction ($H=16$) because short horizon models (e.g., $H=1$) fail to learn anything when not using bootstrapping. For consistency, these models all use regression for outputs that are values and classification for outputs that are collision probabilities.

When not using bootstrapping, models that output values fail to learn, while models that output collision probabilities are extremely sample efficient, stable, and result in high-performing final policies. This dichotomy indicates that learning the dynamics of event probabilities, such as collisions, is easier than learning general, unbounded values. Future work investigating model-based reinforcement learning of domain-specific, discrete events could lead to a new class of sample-efficient model-based algorithms.

When using bootstrapping, models that output values perform worse than models that output collision probabilities. However, models that output values do benefit from using bootstrapping. In contrast, collision prediction models are not strongly affected by using or not using bootstrapping. These results indicate that if the task can be accomplished by looking $H$ steps ahead, then not using bootstrapping can be advantageous.

\begin{figure}[b]
\centering
\vspace*{-20pt}
\includegraphics[width=0.95\columnwidth]{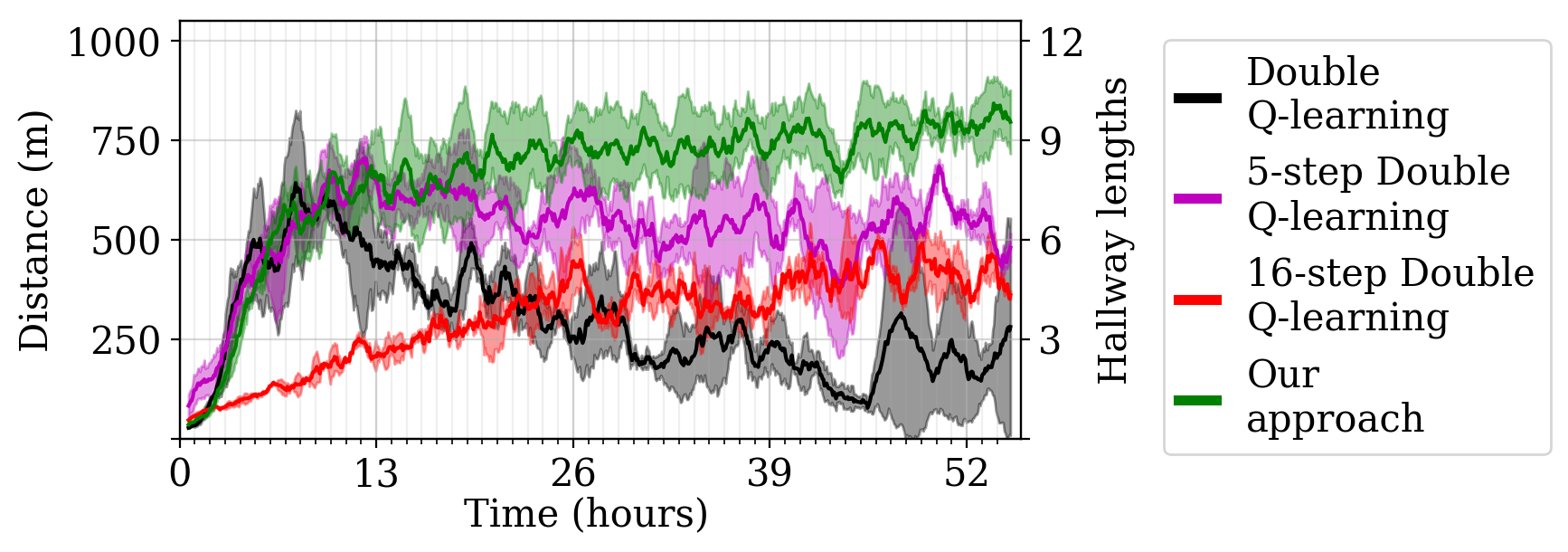}
\caption{Comparison of our robot navigation learning approach to prior methods in a simulated cluttered hallway environment.}
\label{fig:sim-baselines}
\end{figure}

\textbf{Comparisons with prior work.}
Given the empirical evaluations of our design decisions, we choose the instantiation of our generalized computation graph for robot navigation to output collision probabilities, train with a classification loss, use a long model horizon ($H=16$), and not use bootstrapping. To ensure a fair comparison with the prior methods (double Q-learning and $N$-step double Q-learning), we found the best settings for double Q-learning by performing a hyperparameter sweep over relevant parameters, such as exploration rates, learning rates, and target network update rates, and evaluating each set of hyperparameters on a simpler navigation task in an empty hallway. We used these best-performing hyperparameters for all methods in the cluttered hallway environment.

Fig.~\ref{fig:sim-baselines} shows results comparing our approach with double Q-learning and $N$-step double Q-learning; note that we do not compare with model-based approaches because they either assume knowledge of the ground truth state, or the model would have to learn to predict future images, which is sample-inefficient. Our approach is more stable and learns a final policy that is 50\% better than the closest prior method. We believe this highlights that by viewing the problem through the generalized computation graph (Sec.~\ref{sec:gcg}) and incorporating domain knowledge for robot navigation (Sec.~\ref{sec:nav}), we can achieve a sample-efficient, stable, and high performing learning algorithm. 


\subsection{Real-world results}
\label{sec:real-world}

\begin{figure}[b]
\centering
\vspace*{-15pt}
\includegraphics[width=0.8\columnwidth]{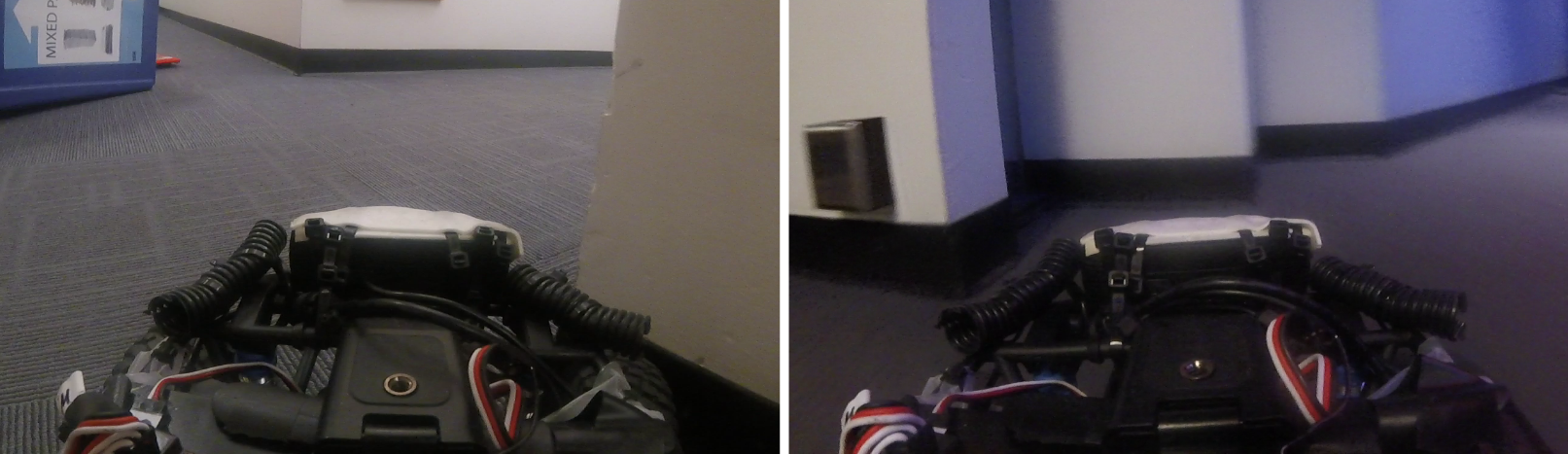}
\caption{Example failure cases in which our approach turns too early (left) and turns too late (right).}
\label{fig:rw-failures}
\end{figure}

We demonstrated the efficacy of our approach on a $\frac{1}{10}$th scale RC car in a real-world environment (Fig.~\ref{fig:teaser}). The task was to navigate the 5th floor of Cory Hall at 1.2m/s, which is challenging as these hallways contain tight turns, changing illumination, and glass walls (Fig.~\ref{fig:rw-pics}).

The RC car learning system was set up to maximize time spent gathering data, minimize computational burden for the car, and be fully autonomous. The computer onboard the RC car is an NVIDIA Jetson TX1, which is intended for embedded deep learning applications. However, all model training is performed offboard on a laptop for computational efficiency, while model inference is performed onboard. We therefore made the system fully asynchronous: the car continuously runs the reinforcement learning algorithm and sends data to the laptop, while the laptop continuously trains the model and periodically sends updated model parameters to the car. For full autonomy, the car automatically detects collisions using the onboard IMU and wheel encoder, and automatically backs up if a collision occurs. The only human intervention required is if the car flips over, which occurred approximately every 30 minutes.

In evaluating our approach, we chose the best design decisions from our simulation experiments: the model outputs are collision probabilities trained using classification, a large model horizon ($H=12$, corresponding to $3.6$m lookahead), and no bootstrapping. All other settings were the exact same as the simulation experiments.

Fig.~\ref{fig:rw-pics} shows that, from training with only 4 hours of data in a complex, real-world environment using only raw camera images and no prior knowledge, the car could navigate significant portions of the environment. For example, the best trajectory travelled 197m, corresponding to nearly 2 loops (and 8 hallway lengths). However, sometimes the policy fails (Fig.~\ref{fig:rw-failures}); additional training should correct these mistakes.

Table~\ref{table:rw-rccar} compares our method with a random policy and double Q-learning trained using the data gathered during our approach's learning. We trained double Q-learning in this way in order to compare the performance of the algorithms given the same state distribution. Our approach travels $17\times$ farther than the random policy and $7\times$ farther than double Q-learning. Qualitatively, our approach was smoothly driving straight when possible, while double Q-learning was exhibiting bang-bang control in that its steering angle is always at the limits.

\begin{table}[t]
\centering
\begin{tabular}{|l|c|c|c|}
\hline \specialcell{Distance until\\crash (m)} & Random policy & \specialcell{Double Q-learning\\with off-policy data} & Our approach \\
\hline Mean                                    & 3.4           & 7.2                                                   & 52.4         \\
\hline Median                                  & 2.8           & 6.1                                                   & 29.3         \\
\hline Max                                     & 8.0           & 21.5                                                  & 197.0        \\
\hline
\end{tabular}
\caption{Evaluation of our learned policy navigating at 1.2m/s using only monocular images in a real-world indoor environment after 4 hours of self-supervised training, compared to a random policy and double Q-learning trained with the same data gathered by our approach.}
\vspace*{-20pt}
\label{table:rw-rccar}
\end{table}


\section{Discussion}
\label{sec:discussion}

We have presented a sample-efficient, stable, and high-performing reinforcement learning algorithm for learning robot navigation policies in a self-supervised manner, requiring minimal human interaction. By formalizing a generalized computation graph that subsumes value-based model-free and model-based learning, we subsequently instantiated this graph to form a suite of hybrid algorithms for robot navigation. Our simulated experiments evaluate which design decisions were important for sample-efficient and stable learning of robot navigation policies, and show our approach outperforms prior Q-learning based methods. Our real-world experiments on an RC car in a complex real-world environment show that our approach can learn to navigate significant portions of the environment using only monocular images with only 4 hours of training in a completely self-supervised manner.

While our approach is able to learn to avoid collisions, the objective was simply to move through the environment at a fixed speed. Future work could investigate how to specify and incorporate higher-level objectives, such as tracking. Also, in these continuous learning environments, exploration is of even greater importance for learning generalizable navigation policies. An intelligent exploration strategy that seek areas of the state space that are novel could increase sample efficiency and policy generalization. Finally, while our approach successfully learned to navigate in a real-world environment, the indoor environment was self-contained and finite. Learning in large outdoor environments with dynamic obstacles presents immense challenges. We believe building robot systems that are robust and can learn using self-supervised approaches will be crucial for getting robots out of laboratories and into the real-world.


\section{Acknowledgements}
\label{sec:ack}

We would like to thank Karl Zipser for the RC car platform and David Gealy for hardware assistance. This research was funded in part by the Army Research Office through the MAST program, the National Science Foundation through IIS-1614653, NVIDIA, and the Berkeley Deep Drive consortium. Gregory Kahn was supported by an NSF Fellowship.


\bibliographystyle{IEEEtran}
\bibliography{2018_ICRA_ProbCollSystem}


\newpage
\clearpage
\appendix

\setcounter{figure}{0}
\renewcommand{\thefigure}{S\arabic{figure}}

\subsection{Additional simulation experiments}

We present additional simulation experimental results to those presented in Sec.~\ref{sec:exp-sim}.

\begin{figure*}
\centering
\includegraphics[width=\textwidth]{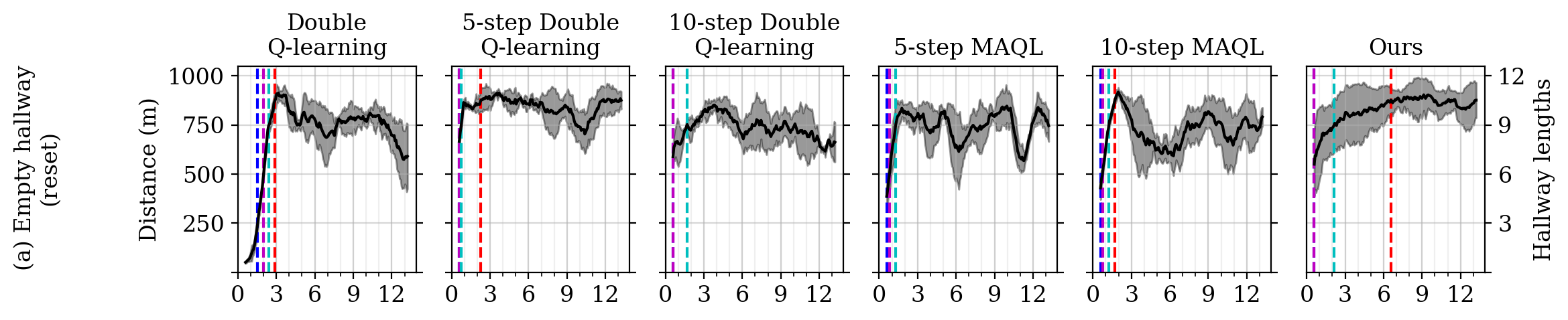}
\includegraphics[width=\textwidth]{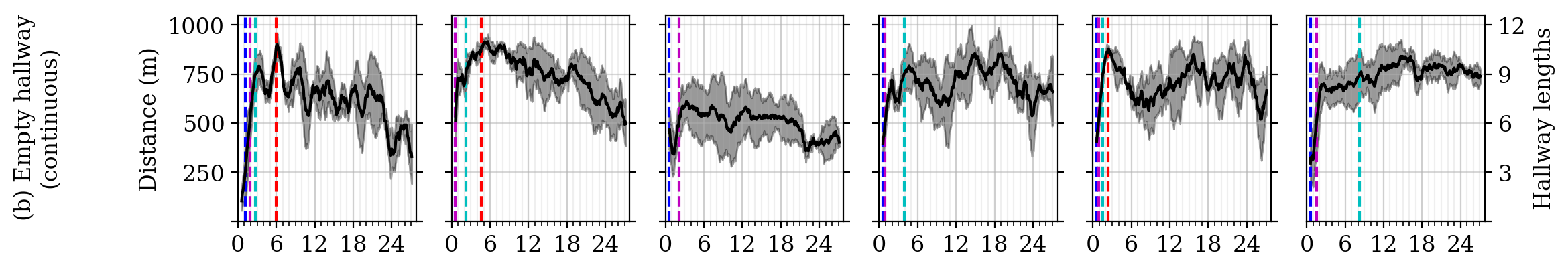}
\includegraphics[width=\textwidth]{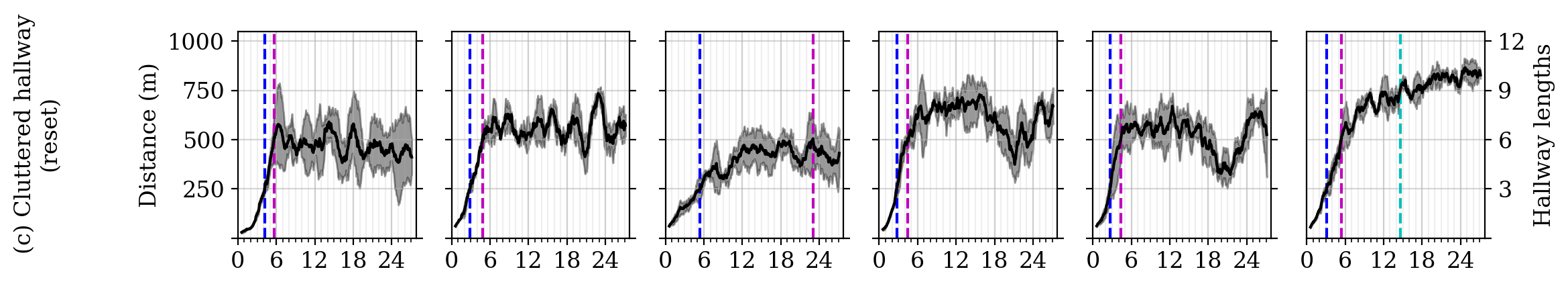}
\includegraphics[width=\textwidth]{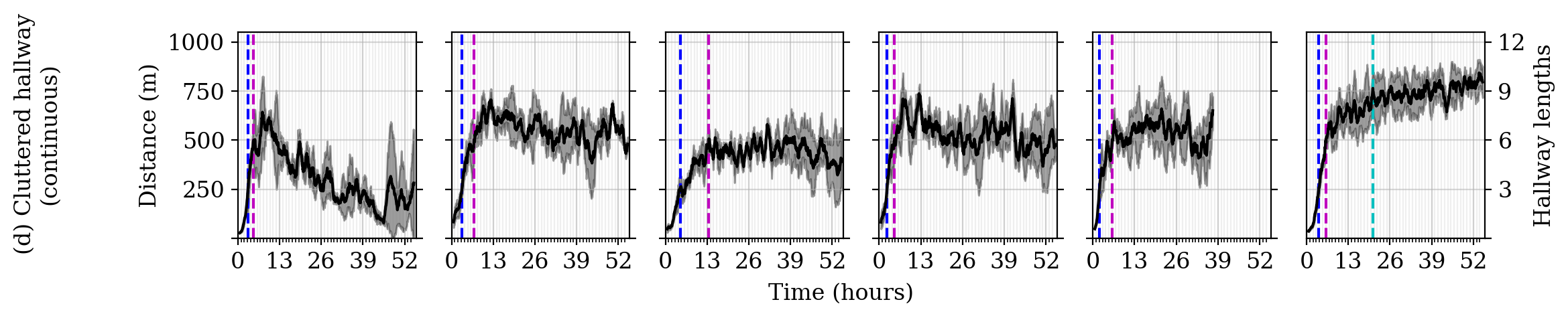}
\caption{Comparison of our approach with double Q-learning, $N$-step double Q-learning, and multi-action Q-learning in indoor environments. The environments were either an empty hallway or a cluttered hallway. For each environment, the car was either reset at the end of each episode, or continuously acted without any resets. Our approach is comparable to the other methods in the empty hallway, but learns faster and is more stable in the complex, cluttered environment.}
\label{fig:appendix-sim-compare}
\end{figure*}

\textbf{Comparisons with prior work}. Fig.~\ref{fig:appendix-sim-compare} shows additional comparisons (beyond Fig.~\ref{fig:sim-baselines}) of our approach to other methods and in different environments. The additional methods we compared to were 10-step double Q-learning (DQL), as well as multi-action Q-learning (MAQL); MAQL is similar to \cite{oh2017value}, in which a multi-action Q-function is used in place of a single-action Q-function.  We considered two environments, an empty hallway and a cluttered hallway. For each environment, the car was either reset at the end of each episode, or continuously acted without any resets.

In the empty hallway with resets (Fig.~\ref{fig:appendix-sim-compare}a), although our approach does learn a high-performing final policy, 5-step DQL learns a similar policy and much quicker. However, when learning continuously (Fig.~\ref{fig:appendix-sim-compare}b), the DQL approaches are unstable, while MAQL and our approach are more stable. In the complex, cluttered environments (Fig.~\ref{fig:appendix-sim-compare}c-d), our approach is the most stable and learns the best performing final policy.

\begin{figure*}
\centering
\includegraphics[width=\textwidth]{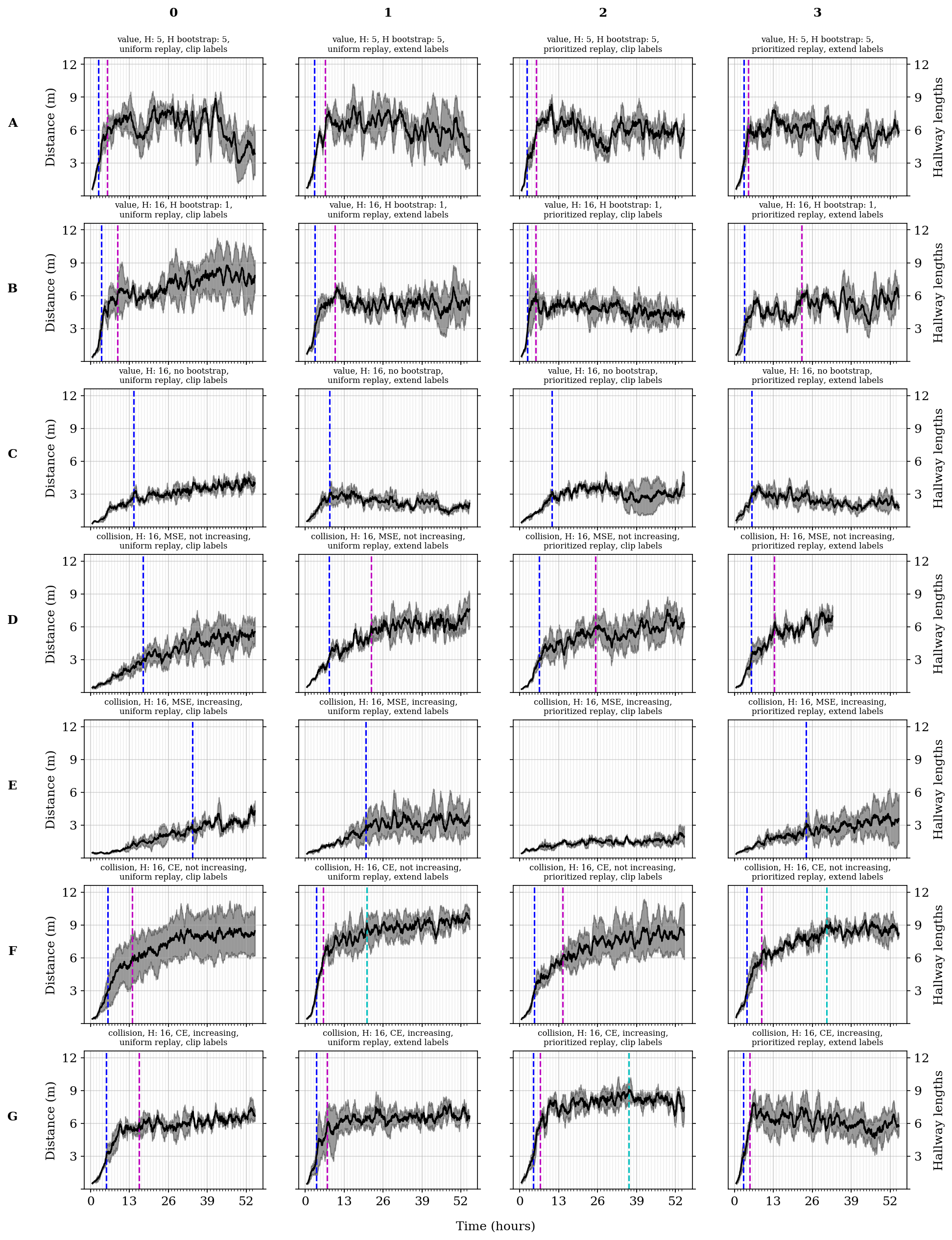}
\caption{Evaluation of the navigation computation graph design decisions in the cluttered hallway environment. The design decisions are: outputting value or collision, model horizon, bootstrapping (and if so, the target network horizon), loss function, enforcing the outputs to be strictly increasing, uniform or prioritized experience replay, and extending or clipping the training labels when an episode terminates. The best design decision we chose for our approach was \textbf{F1}.}
\label{fig:appendix-sim-dd}
\end{figure*}

\textbf{Evaluating design decisions for robot navigation learning}. Fig.~\ref{fig:appendix-sim-dd} contains additional design decision evaluations of our navigation computation graph in the cluttered hallway environment without resets. We evaluated the following design decisions:

\newcommand{\Doutput}{\textbf{D1}}
\newcommand{\Dhorizon}{\textbf{D2}}
\newcommand{\Dbootstrap}{\textbf{D3}}
\newcommand{\Dloss}{\textbf{D4}}
\newcommand{\Dincr}{\textbf{D5}}
\newcommand{\Dreplay}{\textbf{D6}}
\newcommand{\Dclip}{\textbf{D7}}

\begin{enumerate}
\item[\Doutput] Model output (value or collision probability)
\item[\Dhorizon] Model horizon
\item[\Dbootstrap] Bootstrapping
\item[\Dloss] Loss function (mean squared error or cross entropy)
\item[\Dincr] Enforce the model outputs to be strictly increasing
\item[\Dreplay] Uniform versus prioritized experience replay
\item[\Dclip] Extending labels beyond the end of episodes to match the model horizon
\end{enumerate}

In Fig.~\ref{fig:appendix-sim-dd}, the experiments near the top are more similar to MAQL, while the experiments towards the bottom are more similar to our chosen approach. We now analyze each of the design decisions.

\begin{enumerate}
\item[\Doutput] Outputting collision probabilities (rows D-G) is better than outputting expected sum of future rewards (rows A-C) when the other design decisions are chosen appropriately.
\item[\Dhorizon] For MAQL, increasing the horizon from $H=5$ (row A) to $H=16$ (row B) seems to have no effect. Note that for the longer horizon of $H=16$, we had to decrease the horizon of the target network due to computational constraints because training time is roughly  $\mathcal{O}(H^2)$. For outputting collision probabilities, we found $H=16$ was better than $H=12$ (not shown), but we did not investigate horizons longer than 16.
\item[\Dbootstrap] For MAQL, going from using bootstrapping (row B) to no bootstrapping (row C) significantly harms performance. However, for our approach, not using bootstrapping (Fig.~\ref{fig:sim-dd-bootstrapping}) is beneficial.
\item[\Dloss] Using a cross entropy loss (rows F-G) is significantly better than using a mean squared error loss (rows D-E).
\item[\Dincr] When outputting collision probabilities, we can infuse domain knowledge into the model by enforcing the collision probability predictions strictly increase. However, we found that enforcing this increase (rows E and G) was worse than not enforcing this increase (rows D and F).
\item[\Dreplay] We experimented with a form of prioritized experience replay in which experiences ending in collision represented 50\% of each training minibatch. However, we found this form of prioritized experience replay did not help (columns 2-3) compared to uniform experience replay (columns 0-1).
\item[\Dclip] When sampling minibatches from the replay buffer to train the model, some of the samples in the minibatch will be shorter than the model horizon (i.e., $N < H$) because the episode terminated. We have a choice then in how the model is trained: we can clip the labels and only train the model by unrolling the RNN for $N$ steps (columns 0 and 2), or we can extend the labels such that $N=H$ (columns 1 and 3). When extending the labels, the extra timesteps are assumed to consist of random actions and either rewards of zero (in the case of outputting values), or whatever the collision label was at timestep $N$ (in the case of outputting collision probabilities). We found that extending the labels was most often the better design decision. One reason may be that we always perform action selection for $H$ timesteps, so ensuring the model is trained with the same horizon with which it is used is beneficial.
\end{enumerate}

\end{document}